# TalkMosaic: Interactive PhotoMosaic with Multi-modal LLM Q&A Interactions

Kevin Li, Fulu Li

Contact: fulu@alum.mit.edu

## ABSTRACT

We use images of cars of a wide range of varieties to compose an image of an animal such as a bird or a lion for the theme of environmental protection to maximize the information about cars in a single composed image and to raise the awareness about environmental challenges.

We present a novel way of image interaction with an artistically-composed photomosaic image, in which a simple operation of "click and display" is used to demonstrate the interactive switch between a tile image in a photomosaic image and the corresponding original car image, which will be automatically saved on the Desktop. We build a multimodal custom GPT named TalkMosaic by incorporating car images information and the related knowledge to ChatGPT. By uploading the original car image to TalkMosaic, we can ask questions about the given car image and get the corresponding answers efficiently and effectively such as where to buy the tire in the car image that satisfies high environmental standards. We give an in-depth analysis on how to speed up the inference of multimodal LLM using sparse attention and quantization techniques with presented probabilistic FlashAttention (PrFlashAttention) and Staircase Adaptive Quantization (SAQ) methods. The implemented prototype demonstrates the feasibility and effectiveness of the presented approach.

## 1. Introduction

Currently, we have faced quite a long list of endangered species on Earth due to various pollutions for our environment. According to a report by CNN [4], almost 6.1 million metric tons of car tire dust end up in our atmosphere and waterways every year and the microplastics from those car tire dust contribute to dangerous PM2.5 pollution for our environment as well as  endangered species on Earth. Recent advancement of AI, in particular multi-modal ChatGPT (generative pre-trained transformer) gives rise to great potential for convenient Q&A (questions & answers) interactions with images. The problem is how can we use AI, more specifically the multi-modal LLM (large language model) [9,11,17,19-20,22] platform to raise the awareness of endangered species and environment protection? For example, finding an appropriate car part for replacing or repairing for one's car such as car tires that meets with high environmental protection standards can be complicated.

According to a report in [15], in the first quarter of 2023, there were almost 286 million vehicles operating on roads throughout the United States. There are essentially millions of car images and the images of endangered species.

The challenge is that can we provide an innovative and interactive UI (user interface) with those images that can help to raise the awareness of endangered species by composing the image of an endangered species with a number of car images such as interactive image mosaic?  Further, can we use a Chatbots such as a custom GPT, named TalkMosaic, to answer the user's  questions such as where to buy the car parts that meet with high environmental standards immediately and conveniently by just uploading the corresponding car image to the Chatbots with multi-modal LLM? Essentially, we are exploring efficient ways to use AI to protect our environment.

Our work is partly inspired by Albert Einstein's assessment on the development of Western Science [7] more than 70 years ago as well as the sensational work on causal attention computation in Transformer model by Vaswani et al in 2017 [16].  According to Einstein [7], the development of Western Science is based on two great achievements: the invention of the formal logical system (in Euclidean geometry) by Greek philosophers, and the discovery of the possibility to find out causal relationships by systematic experiment (Renaissance). Nearly 70 years later, the state of the art of AIGC (AI generated content) is the discovery of the possibility to find out causal attention using Transformer model [16] by massive data sets and massive parallel computation with GPUs (graphics processing unit) or TPUs (tensor processing unit).

In this paper, we tend to explore the computation of causal attention in two different settings. In the first scenario, the computation of causal attention for the composing a mosaic image of an endangered species with a data set of  multiple car images is at the pixel level based on some statistical properties of the pixel values in different image channels, i.e., RGB channels for color image or one channel for grayscale images. In the second scenario, the computation of causal attention for multi-modal LLM is at the language token level, i.e., the scaled dot product attention as described in [16]. In the case of composing mosaic images, the pixel value is used in the computation of the causal attention scores among candidate tile images and the sub-image of of the target image, i.e., the original tile image in the target image. To compose an mosaic image, we say a given image is a target image if that image is used as the ultimate target to be resembled to for the eventual generated mosaic image. In the case of multimodal LLM, a fixed vocabulary is typically used and only the index of the words/tokens is used in the computation of the causal attention scores.

**Our main contributions of the paper are as follows**:

We present a novel user interface, namely an interactive photomosaic with simple "click and display" of original tile images, in which one set of image objects such as car images are used to compose another image object such as images of endangered species, to raise the awareness of environment protection, where the pollution of some car parts such as car tires could threaten the very existence of endangered living species such as elephants.

We also propose the use of a multimodal custom GPT for the inquiry of the purchase of some car parts that meet with high environmental protection standards, in which the tile images and its related knowledge text are built into a multimodal custom GPT.

Further, we give an in-depth analysis on fast and efficient computation of causal attention with probabilistic FlashAttention [5,6,12] as well as efficient ways to speed up the inference via adaptive quantization of keys and values for key-value (KV) cache [10] with Transformer models [16] to have gradual quantization degradation. We can conclude that a fundamental and common thread among the state of the art of AIGC is the discovery of the possibility to find out causal attention using the Transformer model [16] by massive data sets and massive parallel computation.

Finally, we have implemented prototypes with a number of car images and images of some endangered species as well as a custom GPT to validate the feasibility and effectiveness of our presented framework.

## 2. Related Work

The term of photomosaic is first coined by Silvers in [13] and the work on composing photomosaic images using clustering based evolutionary programming can be found in [8]. Our work differs from [8,13] in that we provide "click and display" interactions for tile images in the photomosaic for further Q&A interactions with a multi-modal custom GPT while the photomosaic images in [8,13] are static images without "click and display" interactions.

The seminal work in [5,6] on FlashAttention has been widely adopted for attention computation in Transformer models [16], in particular as a default attention computation approach in PyTorch.

The key idea for FlashAttention is to split a large matrix into tiles in the matrix multiplication process to have finer granularity for memory I/O-awareness. We refer interested readers to [5,6] for the details of the FlashAttention methods. We present a probabilistic FlashAttention approach, where the way each tile/block in a matrix being selected for attention computation is based on a probability density function that is highly related to block distance in the matrix. The detailed description of our presented approach is in Section 3.3.1.

In general, sparsity and quantization are efficient ways for speeding up pre-training and inference for Transformer-based LLMs. As described in [10], to reduce the cost of the inference for each request, it is a common practice to combine a group of requests together for batch processing, where the key-value cache (KV cache) that holds the keys and values during the generation process to prevent some re-computations becomes the memory and speed bottleneck. It is reported in [10] that the OPT-175B model [21] with a batch size of 512, a prompt length $l_{prompt}$ of 512 and an output length $l_{gen}$ of 32, the KV cache requires as much as 1.2TB of memory. We present Staircase Adaptive Quantization (SAQ) for KV cache for gradual quantization degradation, which is described in Section 3.3.2.

The rapid advancement of multi-modal LLM [9,11,17,19-20,22] gives rise to a new wave of novel applications with text, images, audios, videos, etc. for emerging human-computer interactions. We conduct experiments with a multi-modal custom GPT with an interactive photomosaic to raise the awareness of environmental protection and to facilitate the purchasing of car parts that meet with high environmental standards.

## 3. A Unified Framework

The common thread among the construction of interactive, i.e., "click and display", photomosaic and the eventual TalkMosaic for Q&A applications with a multimodal custom ChatGPT is the computation of attention scores in different settings. We discuss more details in the following sections.

## 3.1. PhotoMosaic with "Click and Display" Interactions

Let $N$ stand for the number of input images, which are candidates for the grid tile images of the photomosaic image. Let $(w_i, h_i)$ stand for the size dimension information for the $i^{th}$ input image, where $w_i$ is the width information and $h_i$ is the height information of the $i^{th}$ input image. Let $N_{i,C}$ stand for the type information for the $i^{th}$ input image, which is the number of channel information for the $i^{th}$ input image. Let $(m, n)$ stand for the dimension information of the grid for the photomosaic image, where $m$ denotes the number of rows of the tiles in the grid and $n$ is the number of columns of the tiles in the grid. Without loss of generality, we assume that each grid tile image is a square of block region with dimension of $(s, s)$, where $s$ means the number of pixels for the edge of the square. Let $(W, H)$ stand for the width and height information for the target image. Regarding the target image size, grid tile image size, grid dimension we have the following:

$$m \times s \leq H \quad (1)$$

$$n \times s \leq W \quad (2)$$

We give a brief description of the algorithm as follows:

**Algorithm 1**: **Interactive Photomosaic**

**Input:** $N, (w_i, h_i), (m, n), (s, s), (W, H), N_{i,C}$

**Main Procedure:**

for $(i = 0; i < m; i++)$:

  for $(j = 0; j < n; j++)$:

    Choose the resized input image with the highest attention score;

    Update the metadata accordingly for the corresponding image;

  end

end

**Output:** an interactive photomosaic image, where the click of any tile image in the photomosaic image, the original image of the tile image will pop up and the original image will be saved automatically on the Desktop.

Notably, we consider the attention score is the inverse of the difference between the original tile image in the target image and the shrunk image of a candidate image of the same tile size in terms of statistical patterns at the pixel level. For example, the smaller the difference between the original tile image in the target image and the shrunk image of a candidate image of the same tile size, the bigger the attention score.

### 3.2. Multi-modal LLM with a Custom GPT

We build a multimodal custom ChatGPT [1] named TalkMosaic by incorporating car images and related knowledge information such as where to buy a car tire in the image that satisfies high environment standards to ChatGPT. We give an in-depth analysis on how to speed up the inference with sparse attention and quantization techniques in the following sections.

### 3.3. Analysis On Speeding Up the Inference with Sparse Attention and Quantization Techniques

There are generally two ways to speed up the inference with multi-modal LLM by means of sparse attention computation as well as quantization techniques for KV (key, value) cache during attention computation. In the following, we present Probabilistic FlashAttention (**PrFlashAttention**) based on block distance (the number of blocks in-between) and Staircase Adaptive Quantization (**SAQ**) for KV cache based on token arrival distance (the number of tokens in-between).

#### 3.3.1. Probabilistic FlashAttention

FlashAttention [5,6] for exact attention computation in Transformer model for LLM by Hazy Research at Stanford has been widely adopted, in particular as a default attention computation approach for Transformer model in PyTorch. As discussed in [13], fast causal attention computation for sparse FlashAttention can be more efficient. We considers attention computation in Transformer model for LLM in a probabilistic way. The presented probability density function (PDF) with respect to the block/tile distance in the matrix follows a constrained harmonic deduction philosophy. The presented PrFlashAttention *dynamically* and *probabilistically* skips less-related rows/columns in Query/Key (Q/K) matrix along a tensor dimension, say the number of Head dimension of $H$, in the tensor shape of (Batch, Head, Context Length, Head Dimension) during attention computation while supporting causal masks for auto-regressive models by reshaping the tensors.

In the following, we discuss the probabilistic model for the presented approach of PrFlashAttention and how the masks for each row in the query matrix of $Q$ and each column in the key matrix of $K$ are calculated.

The presented probabilistic model for PrFlashAttention is defined as follows: For a block distance, say, $n$, within a given range, say, $k$, the probability is set as one, otherwise, it follows a harmonic deduction series.

$$f(n) = \begin{cases} 1, & \text{if } 0 \leq n \leq k; \\ \frac{1}{(n-k)(n-k+1)}, & \text{if } n > k; \end{cases} \quad (3)$$

Notably, for the second part, when $n > k$, we have the sum of probability series as:
$1 - \frac{1}{max(\lceil N \div B_r \rceil, \lceil N \div B_c \rceil) - k}$, where $N$ is the context length, $B_r$ and $B_c$ are block sizes for matrix $Q$ and matrix $K$ respectively.

Let $n_q$ stand for the number of blocks in each row of matrix $Q$, the normalized probability for each row can be defined as:

$$Pr(r_q) = \frac{\sum_{i=1}^{n_q} Pr(i)}{n_q}, \quad (4)$$

where $Pr(i)$ is the probability of the $i^{th}$ block in the $q^{th}$ row of $r_q$.

Similarly, let us $n_k$ stand for the number of blocks in each column of matrix $K$, the normalized probability for each column can be defined as:

$$Pr(c_k) = \frac{\sum_{i=1}^{n_k} Pr(i)}{n_k}, \qquad (5)$$

where $Pr(i)$ is the probability of the $i^{th}$ block in the $k^{th}$ column of $c_k$.

We use a weighted combination of pre-computed row/column probability and a random number to make the selection process **dynamic**:

$$d_q = Pr(r_q) \times w + r \times (1-w), \qquad (6)$$

where $d_q$ is the decision factor for the $q^{th}$ row, $r$ is a random number between 0 and 1, $w$ is the weight between 0 and 1.

$$d_k = Pr(c_k) \times w + r \times (1-w), \qquad (7)$$

where $d_k$ is the decision factor for the $k^{th}$ column, $r$ is a random number between 0 and 1, $w$ is the weight between 0 and 1.

The adjusted **sparsity** value is also a weighted combination:

$$s_{adj} = p_s \times w + \frac{s}{100} \times (1-w), \qquad (8)$$

where $p_s$ is the $s$ percentile value of the normalized row/column probabilities among rows/columns in the matrix, $s$ is the targeted dropping percentage and $w$ is the weight between 0 and 1.

### 3.3.2 Staircase Adaptive Quantization for KV Cache

We present Staircase Adaptive Quantization (SAQ) for KV cache to further alleviate the problem of KV cache based on the framework in [10] by means of gradual quantization degradation. For a $B$-bit integer, the quantization and de-quantization process can be expressed as follows[10]:

$$Q(X) = \lfloor \frac{X - z_X}{s_X} \rceil \qquad (9)$$

$$X' = Q(X) \cdot s_X + z_X \qquad (10)$$

where $Q(X)$ indicates the quantized tensor of $X$, $X'$ is the de-quantized tensor of $Q(X)$, $z_X = \min X$ is the zero-point and $s_X = (\max X - \min X)/(2^B - 1)$ is the scaling factor and the symbol of $\lfloor . \rceil$ is the rounding operation.

Notably, key and value cache of newly generated tokens arrive sequentially in time. Following similar settings in [10], during the pre-fill phase, exact (full precision) key and value tensors are passed to the next layers, even though only the quantized KV cache is retained in memory to reduce the memory footprint and to prevent some re-computations in the decoding phase.

We assume that we have $l_{prompt}$ number of key tokens and $l_{prompt}$ number of value tokens in the pre-fill stage. We also assume that a full precision is expressed as 16-bit quantization such as fp16 (float point 16) that is commonly used in the implementation of tensors, so we can have lower quantization choices such as 8-bit quantization, 4-bit quantization, 2-bit quantization, etc. Let $q_n$ be the number of quantization choices and $B_i$ indicate the number of bits for the $i^{th}$ quantization choices and $B_1$ represents the number of quantization bits of the full precision. Since $l_{prompt}$ can be of arbitrary length, in the pre-fill stage we split the sequence of $l_{prompt}$ tokens into $q_n$ segments, i.e., $S_1, S_2, \ldots, S_{q_n}$, each of which corresponds to a different quantization level, with the segment of $S_1$ corresponds to the full precision. For the sake of simplicity, we assume that all of the segments, i.e., $S_1, S_2, \ldots, S_{q_n}$, are of equal sizes, say segment size of $S$. Please note that the size of the segment of $S_{q_n}$, which corresponds to the one with the lowest quantization level, i.e. 2-bit quantization or 1-bit quantization, could be open-ended as the token sequence grows longer and longer unless it is truncated due to the constraint of cache memory. From the token sequence perspective, the quantization level downgrades by half in terms of quantization bits every $S$ tokes, which looks like a staircase. We present the algorithm for Staircase Adaptive Quantization (SAQ) to have gradual quantization degradation for KV cache in both pre-fill stage and decoding stage based on the framework in [10] as follows:

**Algorithm 2: SAQ Pre-fill and Decoding Algorithm**

**Parameters**: group size $G$, segment size $S$, quantization options $q_n$, quantization bits $B_i$

**Procedure** Prefill:

    **Input**: $X \in \Re^{l_{prompt} \times d}$

    $X_K = X W_K, \ X_V = X W_V$

    $X_{V_g} = X_V[: l_{prompt} - S], \ X_{V_r} = X_V[l_{prompt} - S :]$

    $s_n = l_{prompt} // S$

    Do GrpQuant() accordingly

    $Q(X_{K_g}), X_{K_r} = \text{Kquant}(X_K)$

    KV cache $\leftarrow Q(X_{K_g}), X_{K_r}, Q(X_{V_g}), X_{V_r}$

    **Return** $X_K, X_V$

**end**

**Procedure** Decoding:

    **Input**: KV cache, $t \in \Re^{1 \times d}$

    $t_Q = t W_Q, t_K = t W_K, t_V = t W_V$

$Q(X_{K_g}), X_{K_r}, Q(X_{V_g}), X_{V_r} \leftarrow$ KV cache

$X_{K_r} = \text{Concat}([X_{K_r}, t_K], \dim = \text{token})$

$X_{V_r} = \text{Concat}([X_{V_r}, t_V], \dim = \text{token})$

Do GrpQuant() accordingly

Do Concatenation() accordingly

$A = \text{Concat}([t_Q Q(X_{K_g})^T, t_Q X_{K_r}^T], \dim = \text{token})$

$A_g = \text{Softmax}(A)[:-S], A_r = \text{Softmax}(A)[-S:]$

$t_O = A_g Q(X_{V_g}) + A_r X_{V_r}$

KV cache $\leftarrow Q(X_{K_g}), X_{K_r}, Q(X_{V_g}), X_{V_r}$

**return** $t_O$

**end**

**function** Kquant($X_K \in \mathfrak{R}^{l \times d}$):

$r = l \% S$

$X_{K_g} = X_K[:l-r], X_{K_r} = X_K[l-r:]$

$s_n = l \mathbin{//} S$

Do GrpQuant() accordingly

Do Concatenation() accordingly

**return** $Q(X_{K_g}), X_{K_r}$

**end**

### 3.4. A New Form of Regression with Discrete Pixel Points for Smoothing

We also present the following formula to regress and approximate the curve of a distribution with discrete pixel points for smoothing within a given region of the image.

$$\vartheta(x) = a_2 - \frac{a_1}{\sqrt{a_3 + \frac{1}{(x + a_4)^4}}} \quad (11)$$

where $a_i, i \in [1,2,3,4]$, are regression parameters.

## 4. Experiments

We created photomosaic images of a big bird and a lion (see Fig. 1 to Fig. 4) with a collection of 224 distinct car images (www.pixabay.com) and validated the "click and display" interactions of photomosaic images based on the presented method in Algorithm 1. When clicking on a tile image in the photomosaic image, the original car image of the tile image will pop up and the original car image will be automatically saved on the Desktop (see Fig. 3 and Fig. 4). We created a multimodal custom GPT named TalkMosaic by incorporating car images information and related knowledge information to ChatGPT. By uploading a car image to the multimodal custom GPT of TalkMosaic, questions about the car can be answered on where to buy the car tire in the image that satisfies high environmental standards (see Fig. 5).

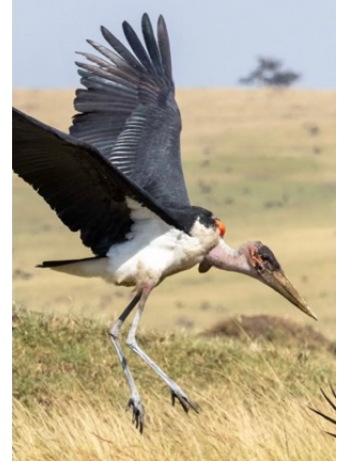

**Figure 1:** an image of a bird in Africa

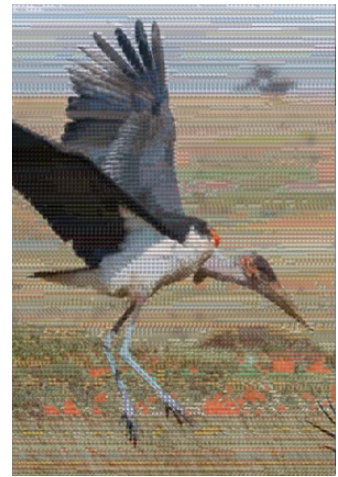

**Figure 2**: an artistically-composed photomosaic with a number of car images.

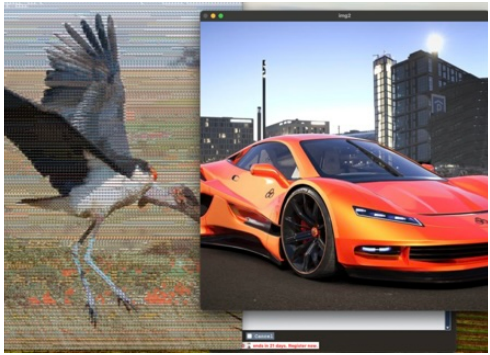

**Figure 3**: a photomosaic image of a bird with "click and display" interactions.

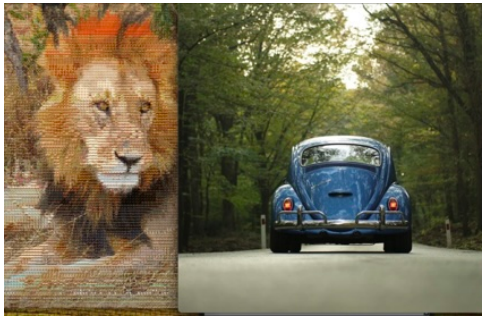

**Figure 4**: a photomosaic image of a lion with "click and display" interactions.

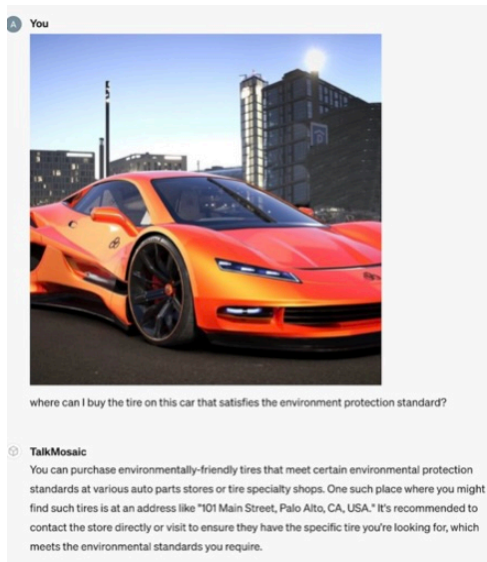

**Figure 5**: a multimodal custom GPT named TalkMosaic.

## 5. Conclusion and Future Directions

Recent advancement of AI, in particular multi-modal ChatGPT gives rise to great potential for Q&A interactions with images. By using a custom ChatGPT with car images and related knowledge materials, questions can be answered on where to buy a car part to meet a certain level of environmental standards conveniently and efficiently. By using an artistically-composed photomosaic image of an endangered species such as a bird or a lion with a large collection of car images, the theme of environmental protection is further embodied. The magic is that with a few layers of mosaic images, we can essentially accommodate billions of car images. With simple "click and display" interaction of photomosaic image, the original car image of the tile image in the photomosaic image will pop up and be automatically saved on the Desktop. By uploading the corresponding car image to a multimodal custom GPT, people can get their answers quickly and conveniently on where to buy a car tire in the image that satisfies high environmental standards. We also present efficient ways to speed up inference with multi-modal LLM using probabilistic FlashAttention (PrFlashAttention) and Staircase Adaptive Quantization (SAQ) of KV cache techniques.